\documentclass{ecai} 

\usepackage{latexsym}
\usepackage{amssymb}
\usepackage{amsmath}
\usepackage{amsthm}
\usepackage{booktabs}
\usepackage{enumitem}
\usepackage{graphicx}
\usepackage{color}

\usepackage{subcaption} 

\usepackage[table,dvipsnames]{xcolor}
\usepackage{tabularx}
\usepackage{multirow}
\definecolor{LightRed}{rgb}{1,0.85,0.85}
\definecolor{LightGreen}{rgb}{0.85,1,0.85}

\usepackage{algorithm}
\usepackage[noend]{algpseudocode}

\usepackage{pifont}

\usepackage{hyperref}

\newtheorem{theorem}{Theorem}

\newcommand{\BibTeX}{B\kern-.05em{\sc i\kern-.025em b}\kern-.08em\TeX}

\begin{document}


\begin{frontmatter}


\paperid{8731} 


\title{
FRAIN to Train: A Fast-and-Reliable Solution\\
for Decentralized Federated Learning
}


\author[A, B]{
    \fnms{Sanghyeon}~\snm{Park}
    \thanks{Email: lukepark@snu.ac.kr.}
}
\author[A]{
    \fnms{Soo-Mook}~\snm{Moon}
    \thanks{Corresponding Author. Email: smoon@snu.ac.kr.}
}

\address[A]{Seoul National University}
\address[B]{Theori Inc.}


\begin{abstract}
Federated learning (FL) enables collaborative model training across distributed clients while preserving data locality.
Although FedAvg pioneered synchronous rounds for global model averaging, slower devices can delay collective progress.
Asynchronous FL (e.g., FedAsync) addresses stragglers by continuously integrating client updates, yet naive implementations risk client drift due to non-IID data and stale contributions.
Some Blockchain-based FL approaches (e.g., BRAIN) employ robust weighting or scoring of updates to resist malicious or misaligned proposals.
However, performance drops can still persist under severe data heterogeneity or high staleness, and synchronization overhead has emerged as a new concern due to its aggregator-free architectures.

We introduce Fast-and-Reliable AI Network, FRAIN, a new asynchronous FL method that mitigates these limitations by incorporating two key ideas.
First, our FastSync strategy eliminates the need to replay past model versions, enabling newcomers and infrequent participants to efficiently approximate the global model.
Second, we adopt spherical linear interpolation (SLERP) when merging parameters, preserving models' directions and alleviating destructive interference from divergent local training.

Experiments with a CNN image-classification model and a Transformer-based language model demonstrate that FRAIN achieves more stable and robust convergence than FedAvg, FedAsync, and BRAIN, especially under harsh environments: non-IID data distributions, networks that experience delays and require frequent re-synchronization, and the presence of malicious nodes.

\end{abstract}

\end{frontmatter}



\begin{table*}[ht]
\footnotesize
\centering
\caption{Comparison of Federated Learning Approaches.}
\label{tab:fl-comparison}
\begin{tabularx}{0.9\textwidth}{l|X|X|X|l}
\hline

\multicolumn{1}{c|}{\textbf{}} &
\multicolumn{1}{c|}{\textbf{FedAvg}} &
\multicolumn{1}{c|}{\textbf{FedAsync}} &
\multicolumn{1}{c|}{\textbf{BRAIN}} &
\multicolumn{1}{c}{{\cellcolor[gray]{0.9}\textbf{FRAIN}}} \\
\hline
\hline

\textbf{Category} &
Federated Learning &
Asynchronous FL &
Blockchain-based FL &
{\cellcolor[gray]{0.9}Blockchain-based FL} \\
\hline

\textbf{Network} &
Synchronous&
Asynchronous&
Decentralized &
{\cellcolor[gray]{0.9}Decentralized + \textsc{FastSync}} \\  
\hline

\textbf{Focus} &
Non-IID data &
Staleness (stragglers) &
Byzantine tolerance &
{\cellcolor[gray]{0.9}Byzantine \& Drift mitigation} \\
\hline

\textbf{Aggregation} &
Data-weighted averaging &
Adaptive $\alpha$ \textsc{Lerp} &
Score-based \textsc{Lerp} &
{\cellcolor[gray]{0.9}Adaptive score-based \textsc{Slerp}} \\
\hline

\end{tabularx}
\bigskip
\end{table*}

\section{Introduction}

Federated Learning (FL) enables multiple clients to collaboratively train a shared global model by leveraging local data on each client without centralizing it at a single location.
A well-known example is \textit{FedAvg}~\cite{fedavg}, which relies on synchronous communication rounds: each client trains the model locally, then sends its updates to a central server that aggregates (averages) them into a global model.
Despite its popularity, FedAvg suffers from the classic straggler problem.
If a subset of clients is slow—either due to limited computational resources, intermittent connectivity, or irregular participation—then the entire global synchronization step is delayed, causing a significant communication bottleneck and lengthening the time to convergence.

Motivated by these limitations, asynchronous FL approaches, such as \textit{FedAsync}~\cite{fedasync}, have attracted growing interest.
In asynchronous settings, each client communicates with the server or network without waiting for others; the global model is updated as soon as a new local update arrives.
By eliminating the need to wait for slow clients, asynchronous FL can, in principle, accelerate learning.
However, a naïve approach that immediately integrates any incoming update is vulnerable to {out-of-date (stale) or misaligned updates}.

Against this backdrop, \textit{BRAIN}~\cite{brain} was proposed to facilitate fully decentralized, asynchronous FL in potentially hostile network environments.
A decentralized network is, by definition, more challenging because there is no trusted central authority---one cannot rely on a single trusted aggregator, and each participant must {verify} every incoming proposal. In BRAIN, to manage the risk of stale or malicious updates, a committee of peers assigns {scores} to each update proposal, reaching consensus on whether (and how) to incorporate it. By doing so {without} any centralized aggregator---who could itself be malicious---BRAIN demonstrated strong {robustness} against both non-IID data distributions and large staleness.
%

Nevertheless, BRAIN still encounters several limitations.
First, by forgoing a centralized aggregator, it must calculate the global model {sequentially} by applying each validated update in turn. As a result, {new participants} or those who re-join after a hiatus face {sync delays}, as they must wait for the model to catch up with all prior updates.
Second, although BRAIN's scoring mechanism mitigates the impact of malicious or stale clients, it can still experience performance degradation when facing extreme {client drift} in highly heterogeneous data scenarios. These issues highlight the potential for further improvement in both {global model efficiency} and {update integration}.

\bigskip

To address these challenges, we propose \textit{FRAIN}, Fast-and-Reliable AI Network, an enhanced asynchronous FL method built on the BRAIN architecture.
%
Our main contributions are as follows:

\begin{itemize}
    \item \textbf{\textsc{FastSync} Strategy.}
    Instead of computing the global model recursively from the initial model by sequentially applying every update, we derive a {pseudo} global model using only the latest proposal and its immediately preceding proposal.
    Consequently, a newly joining node or one re-joining after inactivity can immediately approximate the current global state, eliminating the overhead of replaying older models.
    
    \item \textbf{\textsc{Slerp} for Model Merging.}
    Rather than a plain weighted average of parameters, FRAIN applies Spherical Linear Interpolation (SLERP)~\cite{slerp-q} in the parameter space. SLERP preserves the {direction} property between two parameter vectors, mitigating the destructive interference that can happen when merging significantly diverged local updates.

    \item \textbf{Staleness Penalty Function and \textsc{WiMA}.}
    FRAIN enriches BRAIN’s score-based approach by incorporating {staleness penalty functions} and a {window-based model average} (\textsc{WiMA})~\cite{wima-fl} method.
    The staleness penalty functions reduce the impact of outdated proposals, preventing them from disproportionately influencing the current global model.
    Additionally, FRAIN uses a WiMA-like averaging scheme when computing the mixing coefficient.
    This approach mitigates outliers in a single round and produces smoother (more robust) convergence.
\end{itemize}


Our code is publicly available at \href{https://github.com/BRAIN-chain/FRAIN}{https://github.com/BRAIN-chain/FRAIN} for reproducibility and future research.

\section{Background}
\label{sec:background}

In this section, we provide an overview of federated learning methods, from the classical FedAvg algorithm to asynchronous solutions such as FedAsync, and then discuss BRAIN, which considers fully decentralized networks.
Table~\ref{tab:fl-comparison} provides a high-level comparison of those methods.

We also outline WiMA, a window-based model averaging method that mitigates recent-model bias in federated learning.
Finally, we introduce the basic concept of Spherical Linear Interpolation (SLERP) for model merging.

\subsection{Federated Learning}
\label{subsec:fl}

\paragraph{FedAvg.}
\textit{FederatedAveraging} ({FedAvg})~\cite{fedavg-convergence, fedavg} is a classical {synchronous} Federated Learning (FL) algorithm for aggregating client updates.
At each global round $r$, a central server (aggregator) selects a subset of clients $K_r$ and sends them the current global model $\overline{M_{r-1}}$ (because $\overline{M_{r}}$ will be produced in this round, the $(r-1)$-th global model is the most recent one).
Each client $k \in K_r$ then performs local training using its own dataset, producing a locally updated model $M_r^k$.
The server waits for {all} selected clients to finish training, then receives and aggregates their local models into a new global model:
\begin{equation}
    \label{eq:fedavg}
    \overline{M_{r}} = \sum\nolimits_{k \in K_r} \tfrac{d_k}{d}\, M_r^k
    \nonumber
\end{equation}
where $d_k$ is the number of samples held by client $k$, and $d = \sum_{k \in K_r} d_k$ is the total number of samples among all participating clients in round $r$.
That is, each local update is weighted in proportion to the size of the client’s dataset.

Because FedAvg uses a fully {synchronous} protocol, it can be delayed by {stragglers} (slow clients). In other words, each round’s speed is effectively limited by the slowest client in $K_r$. When clients have heterogeneous computational or network capabilities, FedAvg can experience severe latency, motivating the development of asynchronous solutions.

\paragraph{FedAsync.}
\textit{Asynchronous Federated Optimization}, {FedAsync}, adopts an {asynchronous} structure in which the server updates the global model as soon as it receives a client’s local model~\cite{fedasync}.
When client $k$ finishes local training and sends $M_r$ to the server, the server immediately performs the following update to obtain the global model $\overline{M_r}$:
\begin{equation}
    \label{eq:fedasync}
    \overline{M_{r}} \;=\; (1 - \alpha) \cdot \overline{M_{r-1}} \;+\; \alpha \,M_r \,,
\end{equation}
where $0 \leq \alpha \leq 1$ is a tunable {mixing coefficient}.
In this way, FedAsync effectively eliminates waiting time, potentially accelerating training by utilizing clients as soon as they become available.
Theoretical results show that asynchronous FL can still converge, given certain conditions.

However, such immediate integration can jeopardize {stability} of the global model.
In highly non-IID scenarios or when a client’s local update is based on an {outdated} (stale) global model, the direction of the update can diverge from the global optimum, leading to \textit{client drift}~\cite{client-drift}.
Incorporating such drifted proposals can disrupt stable convergence.
Furthermore, neither FedAvg nor FedAsync inherently defends against malicious clients.
In adversarial settings, an attacker’s model update can resemble a drifted update (significantly different direction), making it difficult to distinguish between harmless stale contributions and genuinely malicious ones.


\subsection{BRAIN}
\label{subsec:brain}

Recognizing that adversarial participants may exist, {BRAIN} (Blockchain-based Reliable AI Network)~\cite{brain} is designed for fully decentralized networks under asynchronous conditions, addressing some of the most challenging real-world scenarios:
Large latency where some participants are significantly delayed;
Potentially adversarial or malicious clients actively joining to disrupt global model convergence;
Highly non-IID data distributions across participants.

\paragraph{Overview.}
In BRAIN, each client trains a local update and proposes it to the network.
Then, the following steps are taken for each proposal:
A committee of peers (training comittee), selected by the smart contract, evaluates the proposal by assigning a {score}, which reflects performance metrics (e.g., accuracy or loss) on their local data.
%
After all individual scores are recorded, the contract derives a consensus value—the median of the submitted scores—to represent reflecting the proposal's reliability.
Let $a_r$ denote the consensus score of the $r$-th proposal $M_r$.
The consensus values recorded in the contract are then used to compute the mixing coefficient $\alpha_r^{(\textsc{BRAIN})}$ as follows:
\begin{equation}
    \label{eq:alpha}
    \alpha_r^{(\textsc{BRAIN})} = \begin{cases}
        \tfrac{a_r}{a_{r-N+1}+\dots+a_{r-1}+a_r} & \text{if $r \ge N-1$}\\
        \tfrac{a_r}{a_0 + \dots + a_{r-1} + a_r} & \text{otherwise},
    \end{cases}
\end{equation}
where $N$ denotes the window size, indicating how many recent consensus scores are aggregated for normalization.
Finally, each client {locally} updates the global model by applying the recent proposal $M_{r}$, weighted by its $\alpha \gets \alpha_r^{(\textsc{BRAIN})}$, via Equation~\ref{eq:fedasync}.
BRAIN can be seen as an {asynchronous extension} of FedAsync with a decentralized scoring mechanism for defending against malicious updates.

\paragraph{Aggregator-Free Synchronization.}
BRAIN removes the centralized aggregator ({aggregator-free}) and replaces its functions with:
\ding{192} A smart contract that stores \textit{scores} assigned by a training committee;
\ding{193} Local clients that use these scores to integrate proposed updates.
By retrieving the same score from the blockchain, all nodes consistently apply the same integration step, thereby maintaining a globally synchronized model without relying on a central aggregator.

However, in BRAIN’s aggregator-free paradigm, the global model must be {sequentially} computed by recursively applying all historical updates in order, which leads to high network overhead.



\subsection{Window-based Model Averaging}
\label{subsec:wima}


\textit{Window-based Model Averaging} (WiMA)~\cite{wima-fl} improves the convergence in federated learning by replacing each global model with the mean of the last $N$ rounds.
This average damps the bias of any single update, smoothing the sharp shifts that non-IID data or long local training can cause, yet introduces no extra communication or significant computation.

As shown in Equation~\ref{eq:alpha}, the mixing coefficient $\alpha$ in BRAIN is also influenced by multiple factors across multiple rounds, and thus results in a damping effect similar to that of WiMA.
Nevertheless, BRAIN and WiMA rest on different principles:
BRAIN assigns the mixing coefficient as the current model’s consensus score divided by the sum of the most recent $N$ scores, whereas WiMA defines it as the simple moving average of those $N$ scores.
It can take a more WiMA-like approach when the $r$-th $\alpha_r^{(\textsc{WiMA})}$ is defined as follows:
%
%
\begin{equation}
    \label{eq:wima_alpha}
    \alpha_r^{(\textsc{WiMA})} = \begin{cases}
         \tfrac{1}{N}
            \sum\nolimits_{k=r-N+1}^{r} a_k & \text{if $r \ge N-1$}\\
         \tfrac{1}{r+1}
            \sum\nolimits_{k=0}^{r} a_k
         & \text{otherwise},
    \end{cases}
\end{equation}


We use WiMA-like method (Equation~\ref{eq:wima_alpha}) in our implementation and experiments instead of BRAIN’s method (Equation~\ref{eq:alpha}) because it yields more stable convergence in practice.
In Section~\ref{subsec:wima-brain-substitutability}, we prove that the two mixing coefficients, $\alpha^{(\textsc{BRAIN})}$ and $\alpha^{(\textsc{WiMA})}$, are mutually substitutable within a bounded error.


\subsection{Spherical Linear Interpolation}
\label{subsec:slerp}

In FedAvg, FedAsync, and BRAIN, the global model update uses a Linear Interpolation (\textsc{Lerp})—Equation~\ref{eq:fedasync} is a straightforward linear combination.
A known issue with pure linear interpolation is that if two vectors have a large angle between them, the resultant vector may have a significantly reduced norm, losing much of its representability.
%
%


Spherical linear interpolation (\textsc{Slerp})~\cite{slerp-q} is commonly used in computer graphics and embedding spaces to interpolate along the shortest arc on the {great circle} connecting two vectors.
Let $\theta$ be the angle between two vectors $\mathbf{v}_1$ and $\mathbf{v}_2$, with
\[
\cos \theta \;=\; \frac{\mathbf{v}_1 \cdot \mathbf{v}_2}{\|\mathbf{v}_1\|\;\|\mathbf{v}_2\|}.
\]
SLERP interpolates between $\mathbf{v}_1$ and $\mathbf{v}_2$ for $0 \le \alpha \le 1$ as:
\begin{equation}
    \label{eq:slerp}
    \textsc{Slerp}(\mathbf{v}_1, \mathbf{v}_2; \alpha) \;=\; 
    \frac{\sin\bigl((1 - \alpha)\,\theta\bigr)}{\sin \theta}\,\mathbf{v}_1
    \;+\;
    \frac{\sin\bigl(\alpha\,\theta\bigr)}{\sin \theta}\,\mathbf{v}_2.
\end{equation}
When $\mathbf{v}_1$ and $\mathbf{v}_2$ differ substantially, SLERP preserves more of each vector’s characteristics (direction and magnitude), thereby avoiding the collapse often seen with LERP.



\section{FRAIN}
\label{sec:frain}

FRAIN builds upon the {asynchronous and decentralized} learning structure of BRAIN but further improves the algorithm to update the global model {faster and more robustly}.
In essence, FRAIN aims to enhance synchronization efficiency and convergence speed in asynchronous federated learning under potentially adverse conditions, such as {non-independent and identically distributed (non-IID) data}, {high latency}, and {Byzantine attacks}.

This section introduces FRAIN’s two key components:
the \textsc{FastSync} strategy, a method to rapidly approximate the latest global model without recursively integrating all historical proposals,
and \textsc{Slerp}‑based model integration, a spherical linear interpolation approach to combining model parameters that mitigates destructive interference caused by directional mismatches.
We also describe various functions for computing the mixing coefficient ($\alpha$), a key parameter shared by both FastSync and SLERP.


\subsection{Overview}
\label{subsec:frain_overview}

\begin{figure}[!t]
    \centering
    \includegraphics[width=1.0\linewidth]{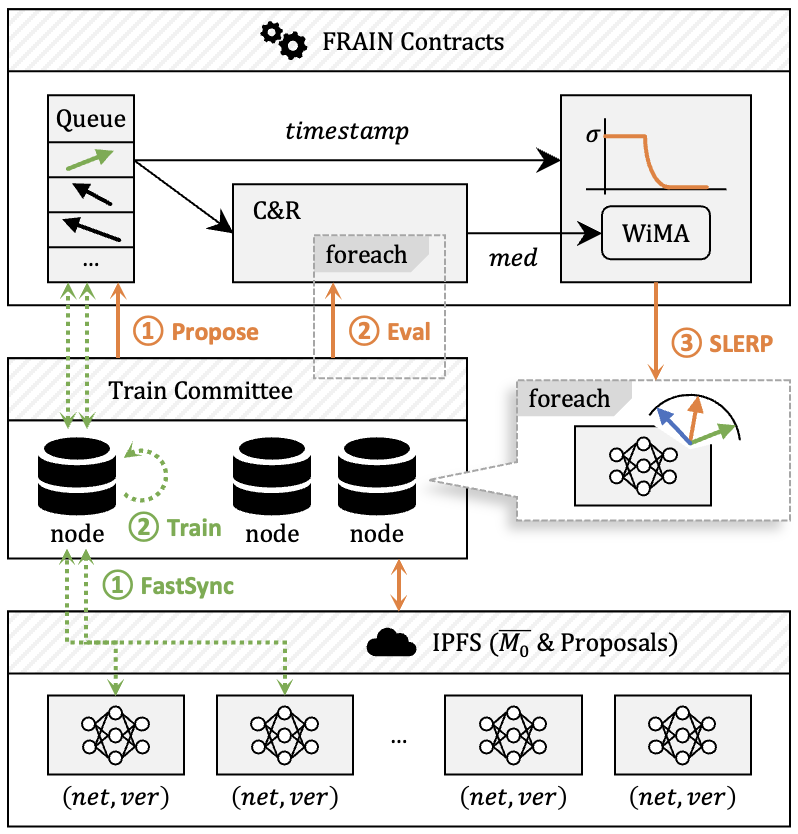}
    \caption{
        Process flow for both \textsc{FastSync} (dotted green) and \textsc{Slerp}-based model merging (orange) in FRAIN.
    }
    \bigskip
    \bigskip
    \label{fig:overview}
\end{figure}

Figure~\ref{fig:overview} illustrates how FRAIN operates in an aggregator‑free decentralized network.

\paragraph{FastSync.}
Each node in training committee retrieves a {pseudo}-global model by aggregating only the latest proposal and its immediately preceding version, rather than replaying the entire history.
With this approximate global model, the node carries out local training on its private dataset.

\paragraph{Model Merging.}
After local training, the node {proposes} its updated model by uploading it (and its version/hash) to IPFS and registering the proposal on the {FRAIN contracts}.  
A set of validators (also known as the {train committee}) then acquires the proposed model, tests its performance locally, and executes a {commit-and-reveal} (C\&R) protocol to upload each validator’s {evaluation score} (e.g., accuracy) to the contract.  
By using C\&R, FRAIN prevents dishonest validators from free-riding on others’ reported scores.

Once all evaluators reveal their scores, the contract computes the {median} and applies a {window-based average} (WiMA-like) of recent medians to derive the mixing coefficient, \(\alpha\).  
A {staleness penalty function}, \(\sigma(\cdot)\), further scales down \(\alpha\) for outdated proposals, thereby discouraging stale updates from dominating the global model.

Finally, each node {locally} merges the new proposal with its current global model using \textsc{Slerp} (Spherical Linear Interpolation), well-preserving the directional information of each update.

\subsection{Fast Synchronization}
\label{subsec:fastsync}

Because BRAIN assumes a fully decentralized network, a newly joining node must {trust only on-chain information} to compute the global model.
%
Specifically, it must use the {initial} global model (whose hash is on-chain), every subsequent proposal model (also on-chain via recorded hashes) in the correct sequence, and the scoring data stored in the smart contract that determines each update’s weight during recursive merging. 
As the number of proposals grows, any node that joins late or remains offline for a while faces the burden of replaying many updates (via Equation~\ref{eq:fedasync}) to recover the {latest} global model.
This not only increases computational overhead but also entails substantial network downloads of model parameters.


FRAIN addresses this {sync delay} problem with the {Fast Synchronization} strategy, \textsc{FastSync}, allowing nodes to approximate the current global model {rapidly} and join training without a full replay of all historical updates.
Instead of recursively applying every past proposal, FastSync uses only the {latest proposal} and its {immediately preceding proposal} to approximate the current global model.
That is, given the two most recent model proposals, $M_{r-1}$ and $M_r$, a new node can quickly infer a pseudo-global model $\widehat{M_r}$ as:
\begin{equation}
    \label{eq:fastsync}
    \widehat{M_r}
    \leftarrow
    \tfrac{\alpha_{r-1} M_{r-1} + \alpha_r M_r}{\alpha_{r-1}+\alpha_r}
    ,
\end{equation}
where $\alpha_{r-1}$ and $\alpha_r$ denote the mixing coefficients for the proposals $M_{r-1}$ and $M_r$, respectively.



Although the approximate global model may be slightly inaccurate, it is typically sufficient to produce local updates whose directions are similar to those based on the exact global model.
This is because:
\begin{itemize}
    \item \textbf{Recent Proposals Dominate.}
    Due to the recursive weighting (via $(1 - \alpha)$ and $\alpha$), older updates quickly lose influence. Thus, two consecutive proposals---especially if both have high scores---already encapsulate most of the effective state of the global model.

    \item \textbf{Latest Includes Prior Contributions.} 
    The latest proposal already incorporates most prior contributions, since it was trained from the then-current global model.

\end{itemize}
Experiments (see Section~\ref{subsec:ablation_fastsync}) show that the \textsc{FastSync}-ed $\widehat{M}$ yields effective approximations, with negligible differences in convergence and final performance relative to the fully recursively integrated $\overline{M}$.


\subsection{SLERP-Based Model Merging}
\label{subsec:slerp_integration}

\begin{figure}[!t]
    \centering
    \begin{subfigure}{.495\linewidth}
        \centering
        \includegraphics[width=\linewidth]{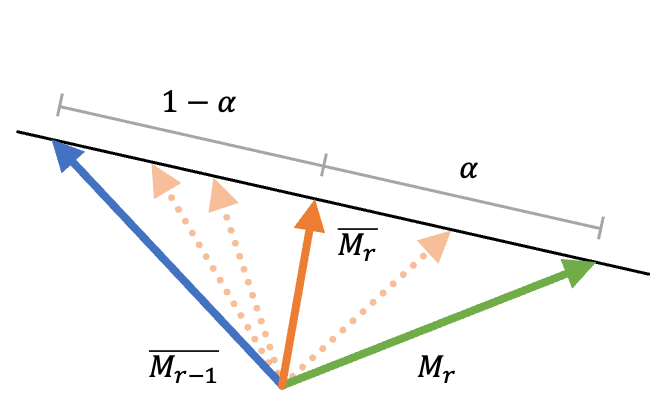}
        \captionsetup{justification=centering}
        \caption{LERP}
        \label{fig:lerp}
    \end{subfigure}
    \hfill
    \begin{subfigure}{.495\linewidth}
        \centering
        \includegraphics[width=\linewidth]{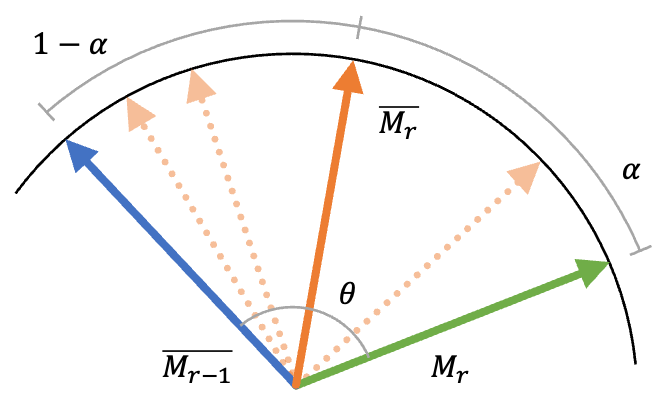}
        \captionsetup{justification=centering}
        \caption{SLERP}
        \label{fig:slerp}
    \end{subfigure}
    \smallskip
    \caption{
        Comparison between LERP and SLERP.
    }
    \bigskip
    \bigskip
    \label{fig:lerp_n_slerp}
\end{figure}

When updating the global model, a simple {linear} blend (\textsc{Lerp}) of weights can suffer if the two model vectors differ sharply in {direction}.
For instance, if two nodes train on significantly different data distributions (non-IID) or if one node works off an old global model (stale), their updates may point in drastically divergent directions.
%
%
In FedAsync and BRAIN, which use LERP (Equation~\ref{eq:fedasync}), a large angle between the global model $\overline{M_{r-1}}$ and the incoming proposal $M_r$ can cause the resulting merged vector—namely, the next global model $\overline{M_r}$—to shrink or to move into a region of parameter space with poor performance.
Figure~\ref{fig:lerp} visually illustrates the decreasing-norm problem in LERP.
%
%

To address this issue, FRAIN adopts {Spherical Linear Interpolation} (\textsc{Slerp}), which blends two vectors along the {great circle} connecting them on a hypersphere. 
SLERP preserves each vector’s magnitude more effectively and prevents the merged model from collapsing into a small‑norm, as shown in Figure~\ref{fig:slerp}.
%
%
%
%
Algorithm~\ref{algo:agg} illustrates how the previous global model and the most recent proposal are aggregated using SLERP, as defined in Equation~\ref{eq:slerp}.

\subsection{Staleness Penalty Functions}
\label{subsec:staleness_function}

Recall that both FastSync and SLERP rely on the mixing coefficient \(\alpha\), which plays a key role by setting how strongly a new proposal influences the synchronization and update of the global model.
%
%
%
In BRAIN, thresholding ensures that proposals with scores below a certain cutoff are effectively ignored.
However,
it does not explicitly address {staleness}.
%
%

From FedAsync, we can find a hint on how to handle staleness well.
FedAsync rescales the mixing coefficient as
\(
    \alpha \gets \alpha\,\times\,\sigma(t - \tau),
\)
where $t$ is the current time, $\tau$ is the timestamp at which the proposal was made, and \(\sigma(t - \tau)\) is a monotonically decreasing function of the staleness \(t - \tau\).
%
%
It introduces three types of weighting function:
\begin{itemize}
    \item \textbf{Constant}: \(\sigma^{(\text{const})}(x)=1\) (no staleness penalty),
    \item \textbf{Polynomial}: \(\sigma^{(\text{poly})}_a(x)=(x+1)^{-a}\),
    \item \textbf{Hinge}: 
\(
    \sigma_{a,b}^{(\text{hinge})}(x)=
        \begin{cases}
            1 &\text{if $x\le b$}\\
            (a(x-b)+1)^{-1} &\text{otherwise}.
        \end{cases}
\)  
\end{itemize}
%
%
The parameters $a$ and $b$ are predefined system constants satisfying $a, b > 0$.

FRAIN extends BRAIN’s scoring scheme by incorporating a FedAsync‑style staleness term; however, deploying this mechanism in a decentralized-and-adversarial setting presents a practical obstacle: the smart contract can rely only on on‑chain data and must remain computationally inexpensive.
Accordingly, we set $\tau$ to the timestamp at which the proposal is enqueued on the smart contract, rather than the potentially unreliable time of its local creation.
We further implement each staleness‑weighting method on‑chain and measure its associated gas cost to evaluate cost‑effectiveness; detailed results are provided in Section~\ref{subsec:ablation_alpha}.



\subsection{WiMA--BRAIN Substitutability}
\label{subsec:wima-brain-substitutability}

To further enhance stability in FRAIN, we replace BRAIN's original update coefficient (which divides the current score $a_r$ by the sum of recent scores) with a {window-based model average} (\textsc{WiMA}) of the last $N$ scores.
Empirically, this helps smooth out any single-round outlier and leads to smoother convergence~\cite{wima-fl}.

However, before doing so, it is crucial to show that switching from BRAIN's ratio-based $\alpha_r^{(\textsc{BRAIN})}$ to WiMA's mean-based $\alpha_r^{(\textsc{WiMA})}$ does not break the convergence properties derived in BRAIN.
%
According to Theorem~\ref{thm:wima-brain-bound}, the difference of two schemes' resulting global models is bounded and does {not} grow unbounded over time;
therefore, WiMA inherits BRAIN's proven guarantees with only a finite offset.

%
%


\begin{theorem}[Bound on WiMA--BRAIN Global Model Difference]
\label{thm:wima-brain-bound}
Let $\overline{M_r}^{(\textsc{BRAIN})}$ and $\overline{M_r}^{(\textsc{WiMA})}$ be global model sequences, both initialized to the same $\overline{M_0}$ yet updated with different mixing weights:
\[
\alpha_r^{(\textsc{BRAIN})}
=
\tfrac{\,a_r\,}{\sum\nolimits_{k=r-N+1}^{\,r} a_k},
\qquad
\alpha_r^{(\textsc{WiMA})}
=
\tfrac{1}{N}\,\sum\nolimits_{k=r-N+1}^{\,r} a_k,
\]
where $a_r \in [\mathcal{T},\,1]$ ($0<\mathcal{T}<1$), and $N\ge1$ is an integer window size.
At round $r$, each method $m \in \{\textsc{WiMA}, \textsc{BRAIN}\}$ updates its global model as:
\[
    \overline{M}_r^{(m)}
    \;=\;
    \bigl(1-\alpha_r^{(m)}\bigr)\,\overline{M}_{r-1}^{(m)}
    \;+\;
    \alpha_r^{(m)}\,M_r.
\]

We assume that each local model $M_r$ is bounded by $\|M_r\|\le B$, and $\|\overline{M_{r}}^{(\cdot)}\|\le B$ as well ($B>0$).
Then, the difference
\[
    \Delta_r
    \;=\;
    \overline{M_r}^{(\textsc{WiMA})}
    \;-\;
    \overline{M_r}^{(\textsc{BRAIN})}
\]
is bounded for every round $r$.
In particular,
\[
    \bigl\|\Delta_r\bigr\|
    \;\le\;
    2B\;\tfrac{\,1 - (1-\mathcal{T})^r\,}{\mathcal{T}}\,
    \quad
    \text{for all $r\ge1$.}
\]
%
As $r \to \infty$, this converges to
\[
    \lim_{r \to \infty}\|\Delta_r\|
    \;\le\;
    \tfrac{2B}{\,\mathcal{T}\,}.
\]
\end{theorem}


\begin{proof}[Proof (Sketch)]
Define
\[
    \Delta_r
    \;:=\;
    \overline{M_r}^{(\textsc{WiMA})}
    \;-\;
    \overline{M_r}^{(\textsc{BRAIN})}.
\]
Using the respective update formulas, we can rearrange it as:
\begin{align*}
    \Delta_r
        &= \overline{M_r}^{(\textsc{WiMA})} - \overline{M_r}^{(\textsc{BRAIN})} \\
        &= (1 - \alpha_r^{(\textsc{WiMA})})\,\overline{M_{r-1}}^{(\textsc{WiMA})} + \alpha_r^{(\textsc{WiMA})}\, M_r \\
            &\qquad - \Big[(1 - \alpha_r^{(\textsc{BRAIN})})\,\overline{M_{r-1}}^{(\textsc{BRAIN})} + \alpha_r^{(\textsc{BRAIN})}\, M_r \Big]\\
        &= (1 - \alpha_r^{(\textsc{WiMA})})\big(\overline{M_{r-1}}^{(\textsc{WiMA})} - \overline{M_{r-1}}^{(\textsc{BRAIN})}\big) \\
            &\quad + \big[(1 - \alpha_r^{(\textsc{WiMA})}) - (1 - \alpha_r^{(\textsc{BRAIN})})\big]\, \overline{M_{r-1}}^{(\textsc{BRAIN})} \\
            &\quad + \big[\alpha_r^{(\textsc{WiMA})} - \alpha_r^{(\textsc{BRAIN})}\big]\, M_r\,\\
        &= (1-\alpha_r^{(\textsc{WiMA})})\,\Delta_{r-1}\\
            &\qquad \;+\; \bigl(\alpha_r^{(\textsc{WiMA})}-\alpha_r^{(\textsc{BRAIN})}\bigr)\,
    \Bigl(M_r - \overline{M_{r-1}}^{(\textsc{BRAIN})}\Bigr).
\end{align*}
%

Taking norms and applying the triangle inequality, we have
\begin{align*}
    \|\Delta_r\|
    &\;\le\;
    (1-\alpha_r^{(\textsc{WiMA})})\,\|\Delta_{r-1}\|\\
        &\qquad \;+\; \bigl|\alpha_r^{(\textsc{WiMA})}-\alpha_r^{(\textsc{BRAIN})}\bigr|\;\cdot\;
    \bigl\|\,M_r - \overline{M_{r-1}}^{(\textsc{BRAIN})}\bigr\|.
\end{align*}
Since $\alpha_r^{(\textsc{WiMA})}\ge \mathcal{T}$, we have $(1-\alpha_r^{(\textsc{WiMA})})\le (1-\mathcal{T})$. 
Meanwhile, $\alpha_r^{(\cdot)}\le 1$, so $\bigl|\alpha_r^{(\textsc{WiMA})}-\alpha_r^{(\textsc{BRAIN})}\bigr|\le 1$. 
Also,
\[
    \|M_r - \overline{M_{r-1}}^{(\textsc{BRAIN})}\|\le \|M_r\|+\|\overline{M_{r-1}}^{(\textsc{BRAIN})}\|\le 2B.
\]
Hence,
\[
    \|\Delta_r\|
    \;\le\;
    (1-\mathcal{T})\,\|\Delta_{r-1}\|
    \;+\;
    2B.
\]
%
%
This is a standard linear recurrence $D_r \le (1-\mathcal{T})\,D_{r-1} + 2B$ with $D_r = \|\Delta_r\|$. In addition, $D_0=0$ since the initial global models are the same for both methods.
Now, iterating yields
\[
    \|\Delta_r\|
    \;\le\;
    2B\,\sum\nolimits_{k=0}^{r-1}\,(1-\mathcal{T})^k
    \;\quad\text{i.e.,}\quad\;
    \|\Delta_r\|
    \;\le\;
    2B\;\tfrac{\,1 - (1-\mathcal{T})^r\,}{\mathcal{T}}\,.
\]
%
As $r\to\infty$, $(1-\mathcal{T})^r\to0$, and the right side converges to $2B/\mathcal{T}$. 
Thus, the difference between WiMA and BRAIN global models is bounded by $2B/\mathcal{T}$ for all time.
\end{proof}


Through these combined mechanisms---SLERP, staleness-weighting function, and WiMA---FRAIN significantly outperforms prior methods in both convergence {stability} and {performance}.
Algorithm~\ref{algo:agg} shows overall model merging method in FRAIN.
\begin{algorithm}[!th]
\caption{Model Merging Using WiMA, Decay, and SLERP}
\begin{algorithmic}[1]

\Require current timestamp $t$, current round $r$,
\Statex predefined decay function $\sigma \in \{ \sigma^{(\text{const})}, \sigma^{(\text{poly})}, \sigma^{(\text{hinge})} \}$,
\Statex window size $N \ge 1$, threshold $\mathcal{T}$ ($0  < \mathcal{T} < 1$),
\Statex scores $\forall a_i \in \{a_1, ..., a_{r-1}\}$ where $\mathcal{T} \le a_i \le 1$, $a_0 = 0$,
\Statex proposal $M_r$ formed at $\tau$ where $\tau \le t$

\Ensure \text{locally calculated $r$-th global model} $\overline{M_r}$

\Procedure{Merge$_r$}{$a_r$, $\overline{M_{r-1}}$, $M_r$}
    \If {$r == 0$} \Return $\overline{M_0} \gets M_0$
    \EndIf
    \smallskip

    \State
    \(
        \alpha_r
        \gets
        \tfrac{1}{\min(N,\,r+1)}
        \sum\nolimits_{k=\max(0,\,r-N+1)}^{r} a_k
    \) \textcolor{blue}{\Comment{\textsc{WiMA}}}

    \State
        \(
            \alpha_r
            \gets
            \alpha_r \times \sigma(t-\tau)
        \) \textcolor{blue}{\Comment{{Decay}}}

    \State \Return
    \(
        \overline{M_r}
        \gets
        \frac{\sin\bigl((1 - \alpha_r)\,\theta\bigr)}{\sin \theta}\,\overline{M_{r-1}}
        \;+\;
        \frac{\sin\bigl(\alpha_r\,\theta\bigr)}{\sin \theta}\,M_r,
    \)
    \State
    \quad \textit{where} \(
        \theta \gets \arccos\!\left( \tfrac{ \overline{M_{r-1}} \cdot M_r }{ \lVert \overline{M_{r-1}} \rVert \,\lVert M_r \rVert } \right).
    \) \textcolor{blue}{\Comment{\textsc{Slerp}}}
\EndProcedure

\end{algorithmic} 
\label{algo:agg}
\end{algorithm}

\section{Experiment}
\label{sec:experiment}

In this section, we evaluate the convergence speed and performance of \textit{FRAIN} against several FL methods: \textit{FedAvg}, \textit{FedAsync}, and \textit{BRAIN}.
We also include \textit{Centralized SGD}, a non‑FL baseline and serves as an upper bound.

\subsection{Experimental Setup}
\label{subsec:exp_setup}

%
Previous studies have focused mainly on CNN-based image classifiers and/or LSTM-based language models, leaving modern Transformer architectures~\cite{transformer} relatively unexplored in federated learning.
To address this gap, we evaluate both a CNN and a Transformer:
\begin{itemize}
    \item \textbf{CNN-based image classifier} trained on CIFAR-10~\cite{cifar-10}.
    We follow the same architecture and hyperparameters used in BRAIN, achieving nearly $94\%$ accuracy under IID centralized training~\cite{94-acc,brain}.
    In this setting, model accuracy serves as the score.
    
    \item \textbf{Transformer-based small language model (sLM)}, specifically the {SmolLM2-135M} model~\cite{smollm2}.
    We reinitialize its weights to investigate how training proceeds from scratch.
    The {inverse of the loss} ($e^{-\,\alpha\,(\text{loss})}$, with $\alpha=0.1$) is used as the score on the WikiText-2~\cite{wikitext-2} validation set.
    In addition, we measure performance via perplexity (PPL) on the WikiText-2 testset.
\end{itemize}

For FedAsync, we adopt a fixed mixing coefficient $\alpha = 0.6$, following its default recommendation.
For BRAIN and FRAIN, we set the moving-average window size to $N=4$ and the score threshold to $\mathcal{T}=0.2$.
In FRAIN, we enable both \textsc{FastSync} and \textsc{Slerp} by default, and we use a {\textit{constant}} decay function for \(\alpha\).
We later assess the individual effects of these components in our ablation studies (Section~\ref{sec:ablation}).
In all FL scenarios, there are 21 nodes in total, and each round (or version update in asynchronous systems) selects 2 nodes to produce local updates.

To reflect realistic scenarios, we simulate {non-IID data distributions}, {communication delays}, and the presence of {Byzantine nodes}.
In the non-IID case, both data quantities and label distributions follow a Pareto distribution~\cite{pareto2,pareto1}.
To simulate staleness, each client’s update is delayed by a uniform random offset of 0--4 (simulated rounds). As a result, some proposals may train on out-of-date (stale) global models.
There are two types of adversaries: \textit{Nullifiers} submit zero-gradient updates thus diluting the global model, while \textit{Randomizers} generate random weights in an attempt to sabotage training.


Each experiment is repeated 10 times; we plot each trial as a dot and the average as a solid line across the ten runs.
All experiments are conducted on a single NVIDIA A100 GPU.

\subsection{Performance and Convergence Speed}
\label{subsec:performance}

\begin{figure}[!t]
    \centering
    \begin{subfigure}{.495\linewidth}
        \centering
        \includegraphics[width=\linewidth]{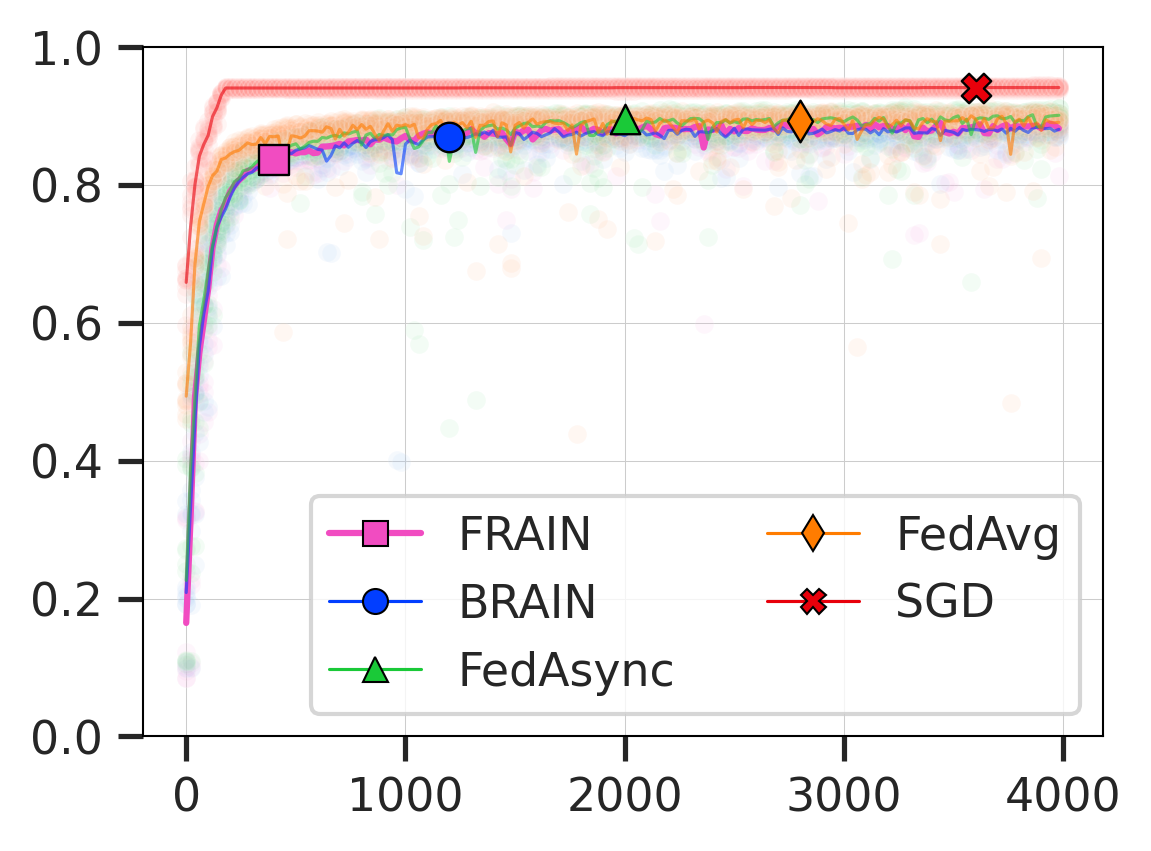}
        \captionsetup{justification=centering}
        \caption{IID}
        \label{fig:convergence-cnn-iid}
    \end{subfigure}
    \hfill
    \begin{subfigure}{.495\linewidth}
        \centering
        \includegraphics[width=\linewidth]{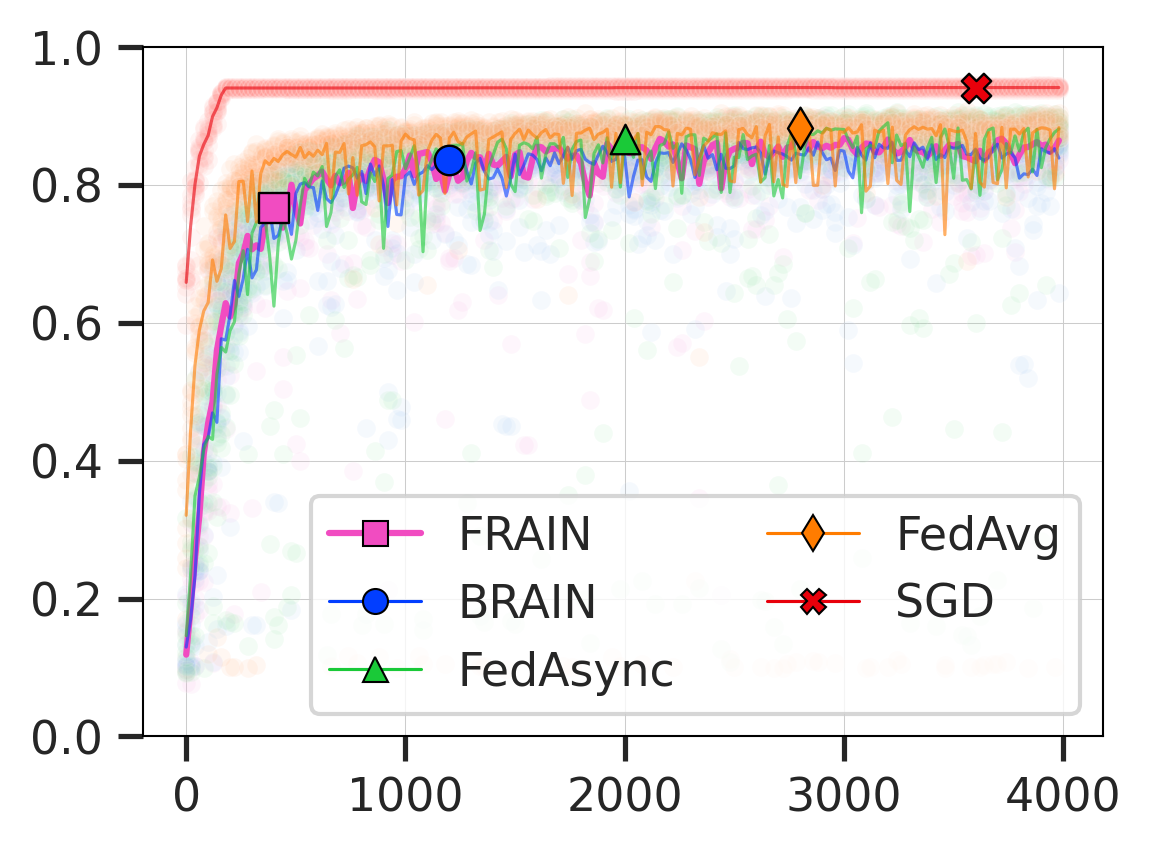}
        \captionsetup{justification=centering}
        \caption{non-IID}
        \label{fig:convergence-cnn-noniid}
    \end{subfigure}
    \smallskip
    \caption{
        Accuracy ($\uparrow$) across gradient updates on CNN/CIFAR-10.
    }
    \bigskip
    \bigskip
    \label{fig:convergence-cnn}
\end{figure}
\begin{figure}[!t]
    \centering
    \begin{subfigure}{.495\linewidth}
        \centering
        \includegraphics[width=\linewidth]{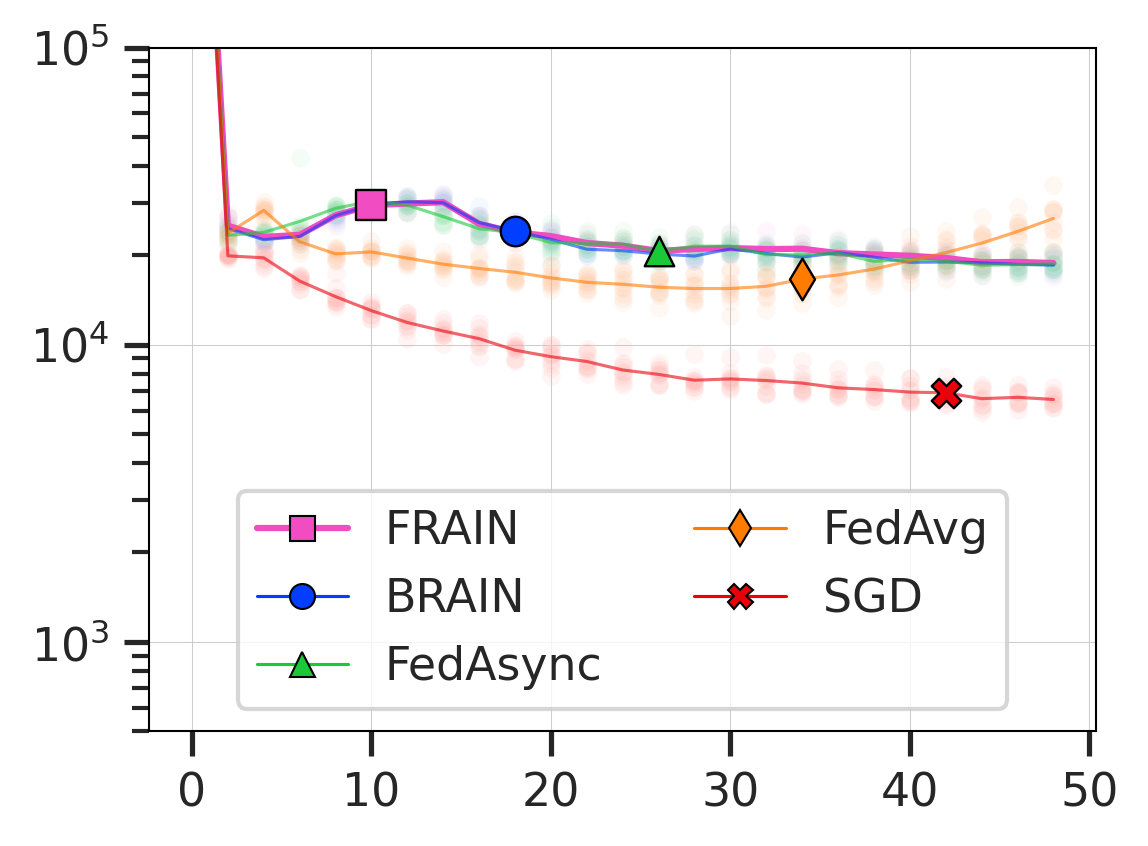}
        \captionsetup{justification=centering}
        \caption{IID}
        \label{fig:convergence-slm-iid}
    \end{subfigure}
    \hfill
    \begin{subfigure}{.495\linewidth}
        \centering
        \includegraphics[width=\linewidth]{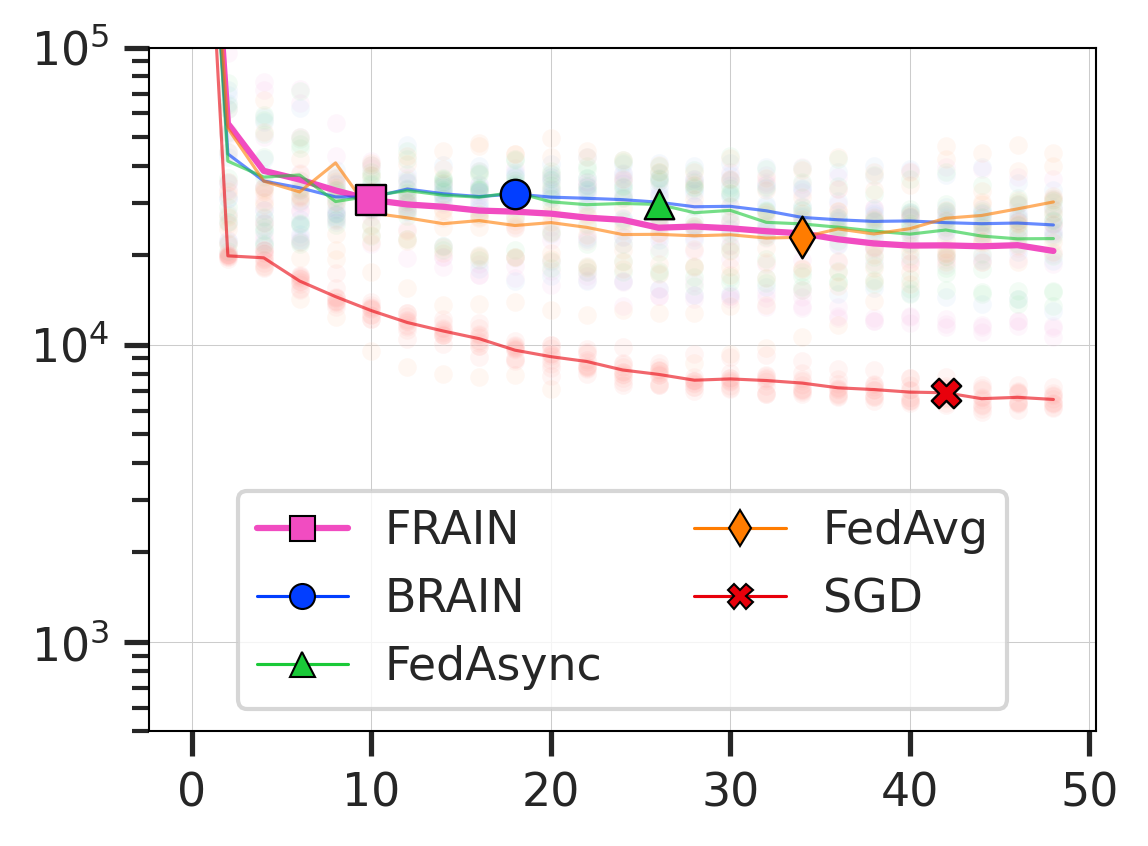}
        \captionsetup{justification=centering}
        \caption{non-IID}
        \label{fig:convergence-slm-noniid}
    \end{subfigure}
    \smallskip
    \caption{
        PPL ($\downarrow$) across gradient updates on sLM-135M/WikiText-2.
    }
    \bigskip
    \bigskip
    \label{fig:convergence-slm}
\end{figure}

We begin by comparing FRAIN against other methods under {IID} and {non-IID} conditions, with no Byzantine participants first. 
%
%
%
%
Figure~\ref{fig:convergence-cnn} presents results for the CNN/CIFAR-10 task, where we plot accuracy (higher is better) as a function of total gradient updates.
In Figure~\ref{fig:convergence-slm}, we show perplexity (lower is better) on the 135M-parameter sLM using the WikiText-2 dataset.
Since we truncated sequences to a maximum length of 1024 and did not apply advanced optimizations (e.g., dataset concatenation), the PPL values may appear relatively high~\cite{ppl-unreliable}. Nonetheless, this setup remains valid for mutual comparing different algorithms.

As a reference, {SGD} serves as the upper bound, since it trains on all data centrally and thus converges fastest.
{FedAvg} also converges quickly because it aggregates every proposed update without filtering.
In contrast, BRAIN and FRAIN filter out low‑scoring proposals. This can slow early convergence, as initial proposals often perform poorly and are discarded; yet the filtering improves stability in more challenging settings (see Section~\ref{subsec:byzantine} for adversarial cases).
In any case, ultimately, all FL methods converge to similar performance under both IID and non-IID conditions.


\subsection{Byzantine Nodes}
\label{subsec:byzantine}

\begin{figure}[!t]
    \centering
    \begin{subfigure}{.495\linewidth}
        \centering
        \includegraphics[width=\linewidth]{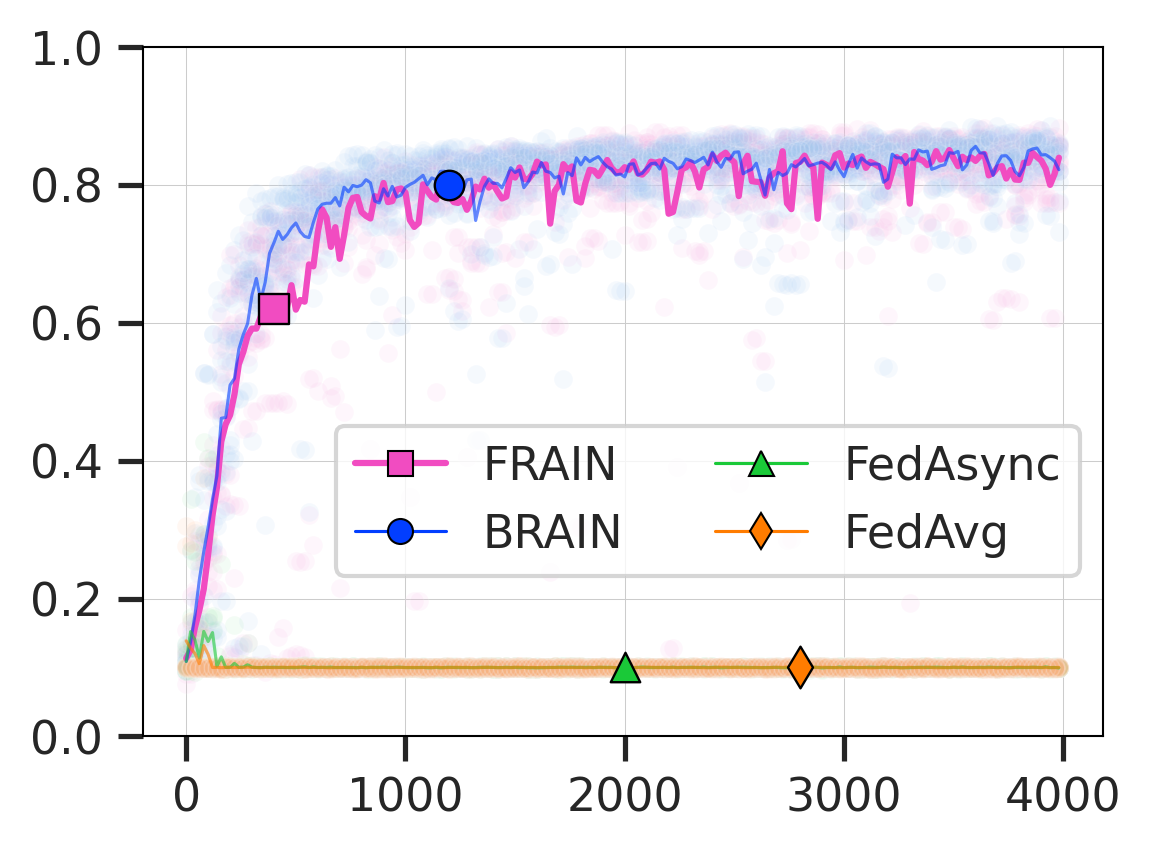}
        \captionsetup{justification=centering}
        \caption{$\text{Nullifiers}=10$}
        \label{fig:byzantine-cnn}
    \end{subfigure}
    \hfill
    \begin{subfigure}{.495\linewidth}
        \centering
        \includegraphics[width=\linewidth]{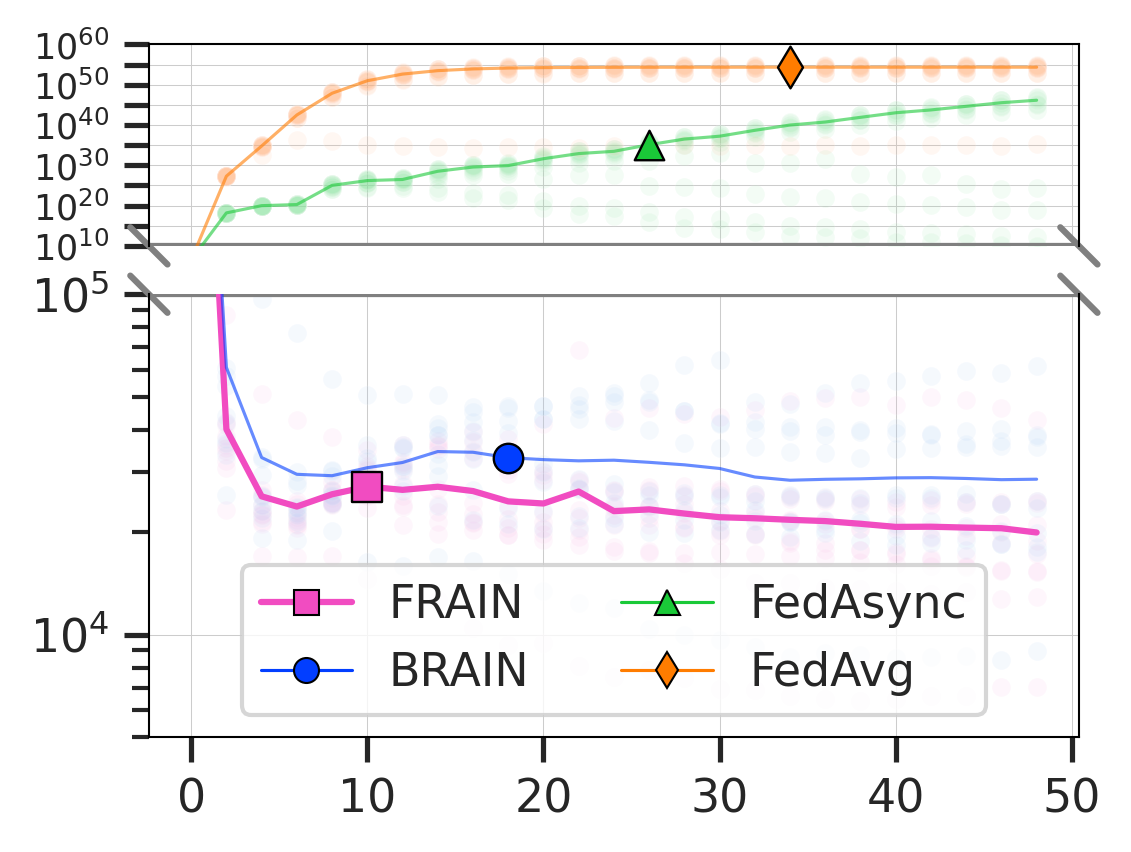}
        \captionsetup{justification=centering}
        \caption{$\text{Randomizers}=10$}
        \label{fig:byzantine-slm}
    \end{subfigure}
    \smallskip
    \caption{
        Gradient updates vs.
        (a) CNN/CIFAR-10 accuracy with nullifiers, and
        (b) sLM-135M/WikiText-2 perplexity with randomizers.
        Both were simulated under a non‑IID environment.
    }
    \bigskip
    \bigskip
    \label{fig:byzantine}
\end{figure}

We now introduce {Byzantine} nodes that either submit nullifying updates or random weights.  
Figures~\ref{fig:byzantine-cnn} and~\ref{fig:byzantine-slm} illustrate the impact of these adversarial strategies on both the CNN (accuracy) and the sLM (perplexity), respectively, under a non-IID environment.
When facing 10 {nullifiers}, both {FedAvg} and {FedAsync} experience drastic performance collapses, sometimes dropping near random-guess accuracy (e.g., $\sim10\%$ on CIFAR-10). 
In the sLM scenario when facing {randomizers}, malicious proposals can drive the model parameters so far off-track that perplexity {explodes}, as visible in Figure~\ref{fig:byzantine-slm}.

{BRAIN} demonstrates considerable robustness thanks to its threshold filtering and median-based scoring.
%
{FRAIN} extends this resilience even further: by combining {WiMA} with {SLERP}, it dilutes unintentionally destructive parameter merges and better suppresses malicious updates.
As a result, FRAIN achieves the strongest robustness under harsh adversarial conditions (especially evident in the sLM/WikiText‑2 task), highlighting the benefits of its {scoring mechanism} and {SLERP-based merging}.

\section{Ablation Studies}
\label{sec:ablation}

We now examine which elements of FRAIN contribute most to its performance gains by removing or modifying specific components.
In particular, we look at the effect of changing the number of nodes that use FastSync, replacing SLERP with standard linear interpolation (LERP), and switching the decaying function of the mixing coefficient (constant, polynomial, or hinge-based).
%

All experiments are conducted under the same configuration described in Section~\ref{sec:experiment}, but we place more emphasis on extreme conditions (e.g., non-IID distributions, networks that require frequent synchronization, and high staleness) to highlight the distinct impact of each design choice.

\subsection{Synchronization}
\label{subsec:ablation_fastsync}

We compare FRAIN \emph{with} vs.\ \emph{without} \textsc{FastSync}.
In the {FastSync} approach, nodes join without calculating a global model, leading to updates that may deviate from the ideal global model.
To evaluate the impact of these drifted proposals, we conducted experiments with various numbers of drifted nodes, relative to a total of 21 nodes.
These nodes do train based on a global-like model calculated through FastSync and then make new proposals.

\begin{figure}[!t]
    \centering
    \begin{subfigure}{.495\linewidth}
        \centering
        \includegraphics[width=\linewidth]{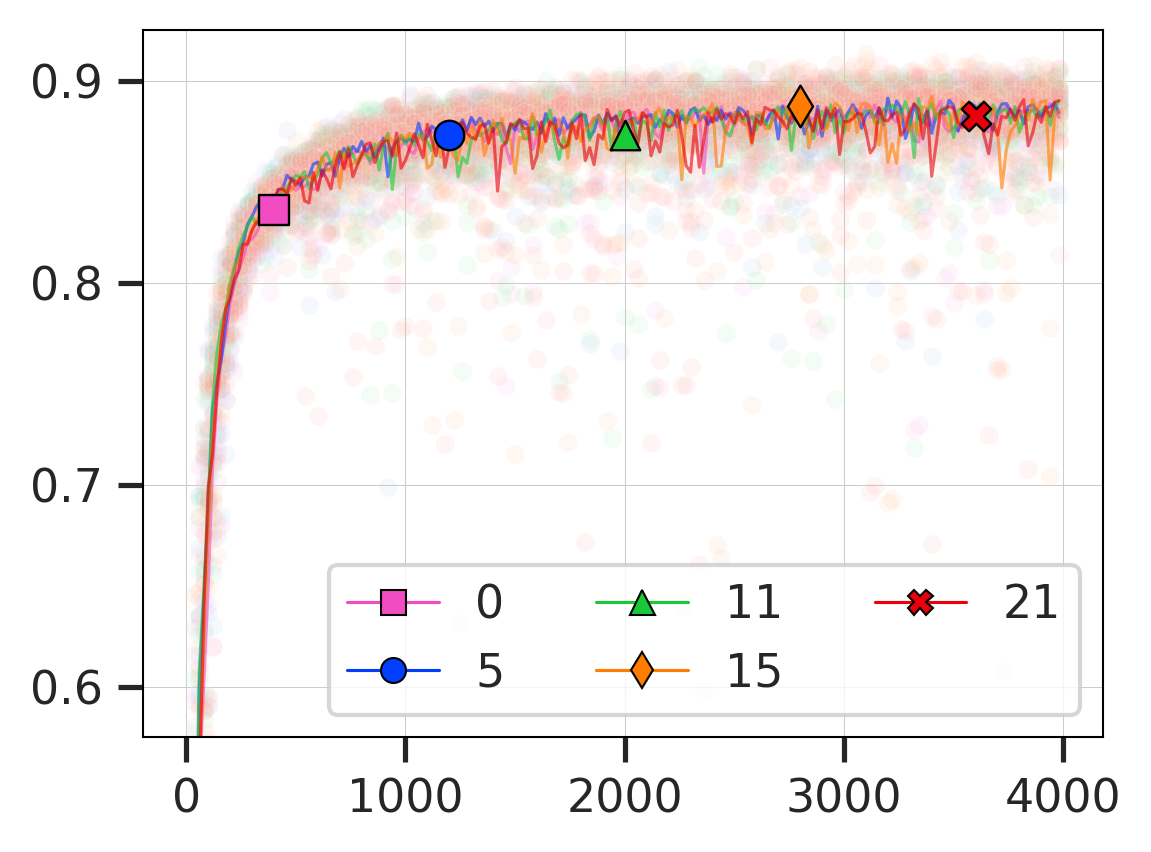}
        \captionsetup{justification=centering}
        \caption{IID}
        \label{fig:fastsync-iid}
    \end{subfigure}
    \hfill
    \begin{subfigure}{.495\linewidth}
        \centering
        \includegraphics[width=\linewidth]{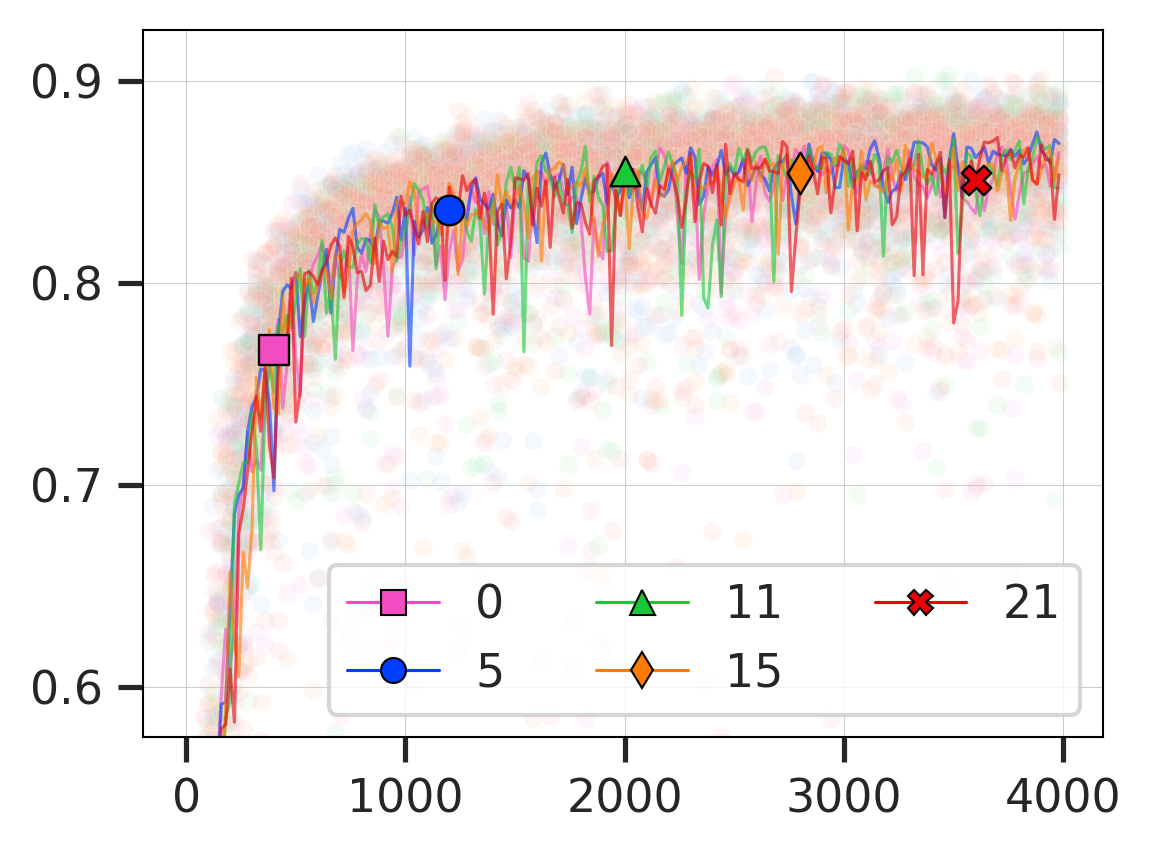}
        \captionsetup{justification=centering}
        \caption{non-IID}
        \label{fig:fastsync-noniid}
    \end{subfigure}
    \smallskip
    \caption{
        Accuracy across gradient updates with varying numbers of \textsc{FastSync}-ed nodes.
    }
    \bigskip
    \bigskip
    \label{fig:fastsync}
\end{figure}

Figure~\ref{fig:fastsync} shows the accuracy on CNN/CIFAR-10 task in IID or non-IID case.
The results show that even with a significant number of drifted nodes, performance remains stable under both IID and non-IID conditions.
Notably, in scenarios where all 21 nodes are drifted, meaning all participating nodes employ the FastSync method without reference to a global model, no performance degradation is observed.

\begin{figure}[!t]
    \centering
    \begin{subfigure}{.495\linewidth}
        \centering
        \includegraphics[width=\linewidth]{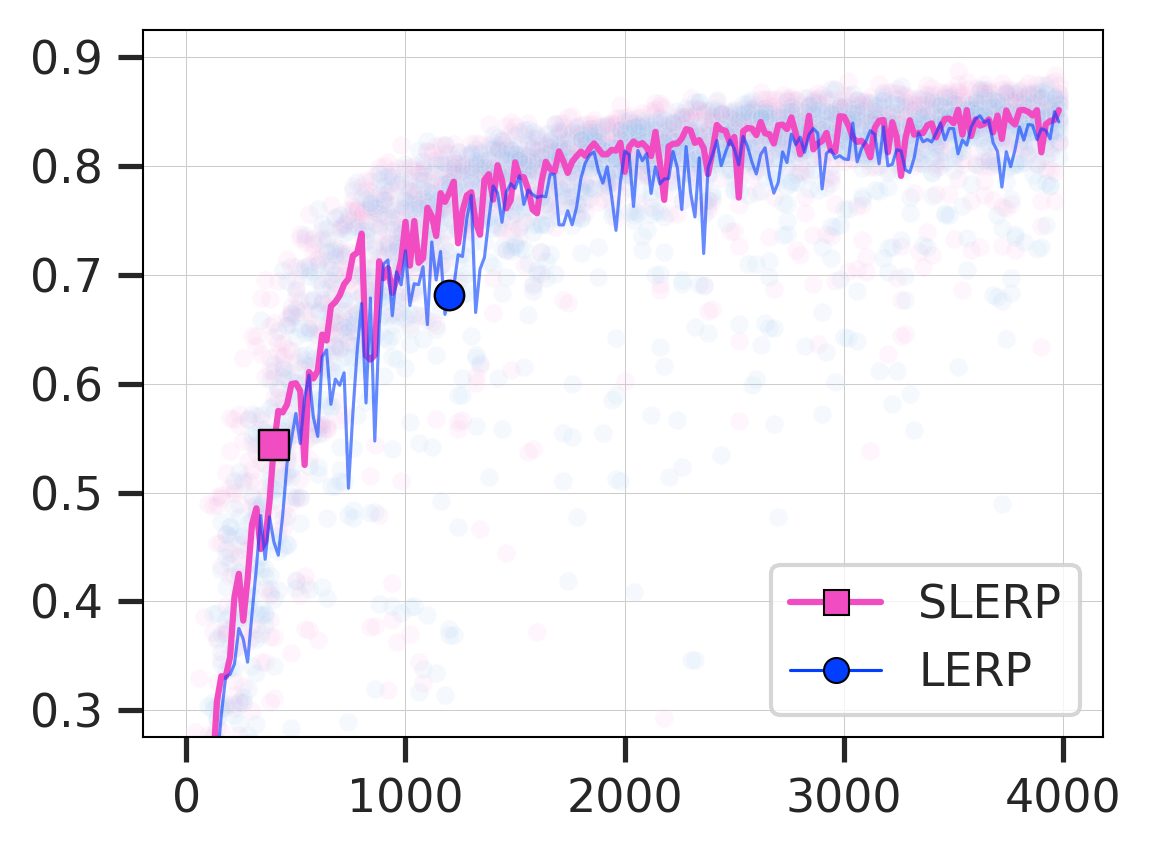}
        \captionsetup{justification=centering}
        \caption{SLERP vs. LERP}
        \label{fig:slerp_n_alpha-slerp}
    \end{subfigure}
    \hfill
    \begin{subfigure}{.495\linewidth}
        \centering
        \includegraphics[width=\linewidth]{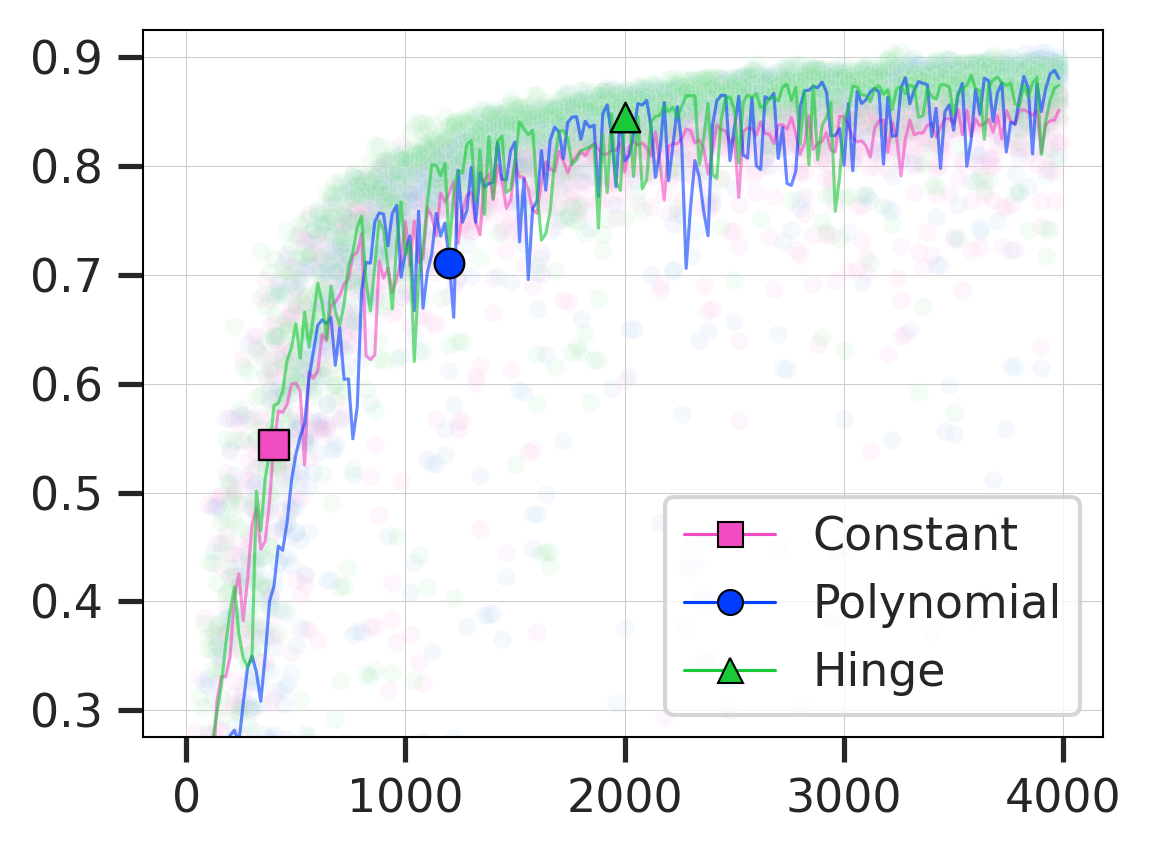}
        \captionsetup{justification=centering}
        \caption{Staleness Functions}
        \label{fig:slerp_n_alpha-alpha}
    \end{subfigure}
    \smallskip
    \caption{
        Accuracy for
        (a) SLERP vs. LERP and
        (b) various staleness-weighting functions,
        simulated under non‑IID with high staleness (max 16) in a setting with $11$ \textsc{FastSync}-ed nodes.
    }
    \bigskip
    \bigskip
    \label{fig:slerp_n_alpha}
\end{figure}

\subsection{SLERP vs.\ LERP}
\label{subsec:ablation_slerp}

We compare SLERP-based merging against a purely LERP-based approach.  
We test this in a harsh scenario: non-IID, maximum staleness of 16 versions, and 11 FastSync nodes.
Figure~\ref{fig:slerp_n_alpha-slerp} shows that SLERP maintains higher accuracy across all rounds under these highly non-IID and stale conditions, by preserving directional consistency in parameter updates.
In addition, SLERP yields a lower standard deviation of about $0.05$, whereas LERP exceeds $0.07$, indicating more stable convergence.

\subsection{Staleness Penalty Strategies}
\label{subsec:ablation_alpha}

We evaluate three staleness penalty strategies in FRAIN: \textit{constant} (no penalty), \textit{polynomial}, and \textit{hinge-based} decays.
Constant method is the simplest to implement but cannot adapt to stale or malicious updates, showing relatively weaker performance under high staleness as illustrated in Figure~\ref{fig:slerp_n_alpha-alpha}.
Polynomial decays handle staleness better than the constant approach, while the hinge-based strategy demonstrates the best performance overall, providing higher accuracy and more stable convergence when dealing with delayed proposals.

\begin{table}[!t]
\centering
\caption{Gas usage benchmark results for score-decaying strategies.}
\label{tab:gas}
    \begin{tabularx}{\columnwidth}{l *{4}{>{\centering\arraybackslash}X}}
    \toprule
        \textbf{Method} & \textbf{Min} & \textbf{Max} & \textbf{Average} & \textbf{Median} \\
    \midrule
        \rowcolor[gray]{0.9}
            \textbf{Constant} & 26365 & 54736 & 53190 & 54736 \\
            \textbf{Poly}     & 36654 & 65025 & 63479 & 65025 \\
            \textbf{Hinge}    & 27527 & 55898 & 54352  & 55898 \\
    \bottomrule
\end{tabularx}
\bigskip
\end{table}

To assess overhead, we measure {gas costs} on the blockchain smart contract for each staleness-decaying scheme.
Table~\ref{tab:gas} reports Ethereum gas usage across 100 tests with random staleness values.
Although polynomial and hinge incur slightly higher costs than constant, the difference remains within a few hundred--thousand gas units, and even hinge-based scoring adds only minimal overhead.
Given the substantial benefits of staleness awareness, the hinge-based approach proves to be both the most effective and the most cost-efficient in practice.


%


\section{Related Work}
\label{sec:related_work}

\paragraph{Client Drift.}
Client drift occurs when local models deviate from the global objective due to non-IID data or stale updates. 
\textit{FedProx}~\cite{fedprox,fedfrox-convergence} adds a proximal term to prevent local updates from straying too far in synchronous FL. 
\textit{SCAFFOLD}~\cite{scaffold} uses control variates to reduce drift, but both methods assume synchronous rounds and do not explicitly handle stale updates.
\textit{FedAsync}~\cite{fedasync} partially addresses staleness in an asynchronous network by adjusting a mixing coefficient \(\alpha\), though it lacks a mechanism to reject malicious proposals.
FRAIN inherits the idea of adjusting \(\alpha\) to mitigate staleness, yet it is fully {asynchronous} and {decentralized}, and it also tackles Byzantine threats.

\paragraph{Non-Linear Interpolation.}
Non-linear parameter merging can reduce loss arising from divergent updates under high heterogeneity or delay~\cite{asyncmanifold,riemannan}. 
For example, \textit{AsyncManifold}~\cite{asyncmanifold} proposes a geometric approach on manifolds to project out conflicting directions.
FRAIN also exploits {spherical} interpolation (\textsc{Slerp}), and thus benefits from robust geometric interpolation.


\section{Conclusion}
\label{sec:conclusion}

In this work, we presented \textit{FRAIN}, an enhanced algorithm building on the \textit{BRAIN} framework for asynchronous federated learning in fully decentralized networks.
FRAIN leverages three key techniques:
\textsc{FastSync} reduces synchronization delays by allowing new or rejoining nodes to quickly approximate the global model, substantially cutting bootstrapping time and communication overhead.
\textsc{Slerp}-based model merging preserves both directional and magnitudinal information during model averaging, mitigating performance degradation caused by client drift and stale updates.
The staleness penalty function and \textsc{WiMA} together smooth out accidental staleness and single-round bias, leading to more stable convergence.
In challenging settings with non-IID distributions, substantial staleness, and even up to $\sim50\%$ Byzantine participants, FRAIN outperforms other FL methods.

\paragraph{Limitations and Future Directions.}
FRAIN currently relies on a set of {static} hyperparameters (e.g., the parameters in the staleness-weighting function).
Although these parameters were chosen empirically, a {dynamic} or {adaptive} mechanism could further enhance performance under highly variable network or data conditions. 
For instance, increasing the staleness penalty when early drift indicators appear, and decreasing it once the global model stabilizes. 
With such {automated strategies}, FRAIN could achieve even faster and more effective merging, making it a promising direction for future research.

\begin{ack}

The work is supported by \href{https://theori.io/}{Theori, Inc.}
\end{ack}



\bibliography{main}

\end{document}